\documentclass[10pt,twocolumn,letterpaper]{article}

\usepackage{iccv}
\usepackage{times}
\usepackage{epsfig}
\usepackage{graphicx}
\usepackage{amsmath}
\usepackage{amssymb}
\usepackage{calc}
\usepackage{subfigure}
\usepackage{amsfonts}
\usepackage{float}
\usepackage{multirow}
\usepackage{color} 

\floatstyle{ruled}
\newfloat{algorithm}{tbp}{loa}
\providecommand{\algorithmname}{Algorithm}
\floatname{algorithm}{\protect\algorithmname}

\usepackage[breaklinks=true,bookmarks=false]{hyperref}

\iccvfinalcopy 

\def\iccvPaperID{424} 

\ificcvfinal\fi
\begin{document}

\title{A Nonparametric Bayesian Approach Toward Stacked Convolutional Independent Component Analysis}

\author{Sotirios P. Chatzis\\
Cyprus University of Technology\\
Limassol 3036, Cyprus\\
{\tt\small sotirios.chatzis@eecei.cut.ac.cy}
\and
Dimitrios Kosmopoulos\\
University of Patras\\
Agrinion 30100, Greece\\
{\tt\small dkosmo@upatras.gr}
}

\maketitle

\begin{abstract}
Unsupervised feature learning algorithms based on convolutional formulations
of independent components analysis (ICA) have been demonstrated to
yield state-of-the-art results in several action recognition benchmarks.
However, existing approaches do not allow for the number of latent
components (features) to be automatically inferred from the data in
an unsupervised manner. This is a significant disadvantage of the
state-of-the-art, as it results in considerable burden imposed on
researchers and practitioners, who must resort to tedious cross-validation
procedures to obtain the optimal number of latent features. To resolve
these issues, in this paper we introduce a convolutional nonparametric
Bayesian sparse ICA architecture for overcomplete feature learning
from high-dimensional data. Our method utilizes an Indian buffet process
prior to facilitate inference of the appropriate number of latent
features under a hybrid variational inference algorithm, scalable
to massive datasets. As we show, our model can be naturally used to
obtain deep unsupervised hierarchical feature extractors, by greedily
stacking successive model layers, similar to existing approaches.
In addition, inference for this model is completely heuristics-free;
thus, it obviates the need of tedious parameter tuning, which is a
major challenge most deep learning approaches are faced with. We evaluate
our method on several action recognition benchmarks, and exhibit its
advantages over the state-of-the-art.
\end{abstract}

\section{Introduction}

Unsupervised feature learning from high-dimensional data using deep
sparse feature extractors has been shown to yield state-of-the-art
performance in a number of benchmark datasets. The major advantage
of these approaches consists in the fact that they alleviate the need
of manually tuning feature design each time we consider a different
sensor modality, contrary to conventional approaches, such as optical
flow-based ones (e.g., HOG3D \cite{hog3d}, HOG/HOF \cite{hoghof}),
methods maximizing saliency functions in the spatiotemporal domain
(e.g., Cuboid \cite{Cuboid} and ESURF \cite{esurf}), methods based
on dense trajectory sampling (e.g., \cite{HengWang2011} and \cite{pipeline}),
and methods based on hierarchical template matching after Gabor filtering
and max pooling (e.g., \cite{jhuang}).

Indeed, unsupervised feature extractors are highly generalizable,
being capable of seamlessly learning effective feature representations
from observed data, irrespectively of the data nature and/or origin.
Due to this fact, there is a growing interest in such methods from
the computer vision and machine learning communities, with characteristic
approaches including sparse coding (SC) \cite{sc,sc2}, deep belief
networks (DBNs) \cite{dbn}, stacked autoencoders (SAEs) \cite{sae},
and methods based on independent component analysis (ICA) and its
variants (e.g., ISA \cite{isa} and RICA \cite{rica}).

In this work, we focus on unsupervised feature extractors based on
stacked convolutional ICA architectures, with application to action
recognition in video sequences. Our interest in these methods is motivated
by both experimental results, where such approaches have been shown
to yield state-of-the-art performance, as well as results from neuroscience,
where it has been shown that these algorithms can learn receptive
fields similar to the V1 area of visual cortex when applied to static
images and the MT area of visual cortex when applied to sequences
of images \cite{neurosc,neurosc2,neurosc3}. 

A major drawback of existing deep learning architectures for feature
extraction concerns the requirement of \emph{a priori }provision of
the number of extracted latent features \cite{cdbn,cite1,cite2,cite3,isa}.
This need imposes considerable burden to researchers and practitioners,
as it entails training multiple alternative model configurations to
choose from, and application of cross-validation to determine optimal
model configuration for the applications at hand. Therefore, enabling
automatic data-driven determination of the most appropriate number
of latent features would represent a significant leap forward in the
field of deep unsupervised feature extraction approaches. 

To address these issues, in this paper we initially introduce a nonparametric
Bayesian sparse formulation of ICA. Our model imposes an Indian buffet
process (IBP) \cite{ibp,ibp2} prior over the learned latent feature
matrix parameters, that naturally promotes sparsity, and allows for\emph{
automatically inferring the optimal number of latent features}. We
underline that the IBP prior is designed under the assumption of infinite-dimensional
latent feature representations, thus being capable of \emph{naturally
handling extraction of overcomplete representations} if the data requires
it, without suffering from degeneracies \cite{ibp-sb-vb}\footnote{As an aside, we also note that performing Bayesian inference over
the parameters of ICA-based models (instead of the point-estimates
obtained by existing approaches) allows for taking \emph{uncertainty}
into account during the learning procedure \cite{vbg}. Even though
this is not examined in this paper, such a capacity is theoretically
expected to yield much better performing models in cases learning
is conducted using \emph{limited }and \emph{scarce }datasets \cite{vbtmfa}.}. We dub the so-obtained model as \emph{IBP-ICA}.

We devise an efficient inference algorithm for our model under a \emph{hybrid
variational }inference paradigm, similar to \cite{hvb}. In contrast
to traditional variational inference algorithms, which require imposition
of truncation thresholds for the model or the variational distribution
over the extracted features \cite{vbg}, our method adapts model complexity
on the fly. In addition, variational inference scales much better
to massive datasets compared to Markov chain Monte Carlo (MCMC) approaches
\cite[Chapter 10]{BishopBook}, which do not easily scale, unless
one resorts to expensive parallel hardware. 

Finally, we apply our IBP-ICA model to the problem of action recognition
in video sequences. For this purpose, we present a stacked convolutional
architecture for unsupervised feature extraction from data with spatiotemporal
dynamics, utilizing IBP-ICA models as its building blocks. Our convolutional
architecture, hereafter referred to as \emph{stacked convolutional
IBP-ICA} (SC-IBP-ICA) networks, is inspired 
from related work on convolutional neural networks, e.g. 
the 3D-CNN method \cite{3dcnn}, and is based on the approach 
followed by existing convolutional extensions of unsupervised
feature extractors, e.g. convGRBM \cite{grbm} and ISA \cite{isa}.
Specifically, similar to \cite{grbm,isa}, our convolutional architecture
comprises training one unsupervised feature extractor (in our case,
one IBP-ICA model) on small spatiotemporal patches extracted from
sequences of video frames, and subsequently convolving this model
with a larger region of the video frames. Eventually, we combine the
responses of the convolution step into a single feature vector, which
is further processed by a pooling sublayer, to allow for translational
invariance. The so-obtained feature vectors may be further presented
to a similar subsequent processing layer, thus eventually obtaining
a deep learning architecture. 

Our stacked model is greedily trained in a layerwise manner, similar
to a large number of alternative approaches proposed in the deep learning
literature \cite{dbn,cdbn,sae}. Our hybrid variational inference
algorithm for this model is completely \emph{heuristic parameter-free,}
thus obviating the need of parameter tuning, which is a major challenge
most deep unsupervised feature extractors are faced with.

The remainder of this paper is organized as follows: In Section 2,
we briefly review existing ICA formulations for unsupervised feature
extraction. In Section 3, we present our proposed method, and elaborate
on its inference and feature generation algorithms. In Section 4,
we experimentally demonstrate the advantages of the proposed approach:
we apply it to the Hollywood2, YouTube, and KTH action recognition
benchmarks. Finally, in the last section we summarize our results
and conclude this paper.

\section{ICA-based feature extractors}

In this section, we provide an overview of existing ICA-based feature
extractors, which are relevant to our approach. Let us denote as $\{\boldsymbol{x}_{n}\}_{n=1}^{N}$
a random sample of size $N$ comprising $D$-dimensional observations.
ICA, in its simplest form, models the observed variables $\boldsymbol{x}_{n},n=1,...,N,$
as 
\begin{equation}
\boldsymbol{x}_{n}=\boldsymbol{G}\boldsymbol{y}_{n}+\boldsymbol{e}_{n}
\end{equation}
where $\boldsymbol{y}_{n}$ is a $K$-dimensional vector of latent
variables (latent features), $\boldsymbol{G}$ is a $D\times K$ matrix
of factor loadings (latent feature matrix), and $\boldsymbol{e}_{n}$
is the model error pertaining to modeling of $\boldsymbol{x}_{n}$.
ICA assumes that $(\boldsymbol{x}_{1},\boldsymbol{y}_{1}),(\boldsymbol{x}_{2},\boldsymbol{y}_{2}),...,(\boldsymbol{x}_{N},\boldsymbol{y}_{N})$
are independent, identically distributed (i.i.d). Further, the \emph{key}
characteristic of ICA that sets it apart from related approaches is
the additional assumption that the distinct components (features)
comprising the feature vectors $\boldsymbol{y}_{n}=[y_{nk}]_{k=1}^{K}$
are also i.i.d. For example, we may consider a simple $J$-component
mixture of Gaussians (MoG) prior, i.e.
\begin{equation}
y_{nk}\sim\sum_{j=1}^{J}\varpi_{kj}\mathcal{N}(0,s_{kj})
\end{equation}
with a Dirichlet prior imposed over the weight vectors $\boldsymbol{\varpi}_{k}=[\varpi_{kj}]_{j=1}^{J}$,
i.e.
\begin{equation}
p(\boldsymbol{\varpi}_{k}|\boldsymbol{\xi}_{k})=\mathrm{Dir}(\boldsymbol{\varpi}_{k}|\boldsymbol{\xi}_{k})
\end{equation}
 and a Gamma prior imposed over the inverse variances $s_{kj}^{-1}$
\begin{equation}
p(s_{kj}^{-1})=\mathcal{G}(s_{kj}^{-1}|\eta_{1},\eta_{2})
\end{equation}
 Finally, model error is usually considered to follow an isotropic
Gaussian distribution, reading
\begin{equation}
\boldsymbol{e}_{n}\sim\mathcal{N}(0,\phi^{-1}\boldsymbol{I})
\end{equation}

Along these lines, several researchers have also considered more complex
assumptions regarding the model likelihood expression. For instance,
a \emph{nonlinear likelihood }assumption has been adopted in \cite{isa},
yielding $\boldsymbol{x}_{n}\approx\sigma(\boldsymbol{G}\boldsymbol{y}_{n})$,
where $\sigma$ is some nonlinear function (e.g., quadratic). Eventually,
the training algorithm of the model reduces to a minimization problem
that takes the form
\begin{equation}
\underset{\boldsymbol{G}}{\mathrm{min}}\sum_{k=1}^{K}\sum_{n=1}^{N}h(\boldsymbol{G}_{:,k}^{T}\boldsymbol{x}_{n})
\end{equation}
where the form of the function $h(\cdot)$ follows from the form of
the postulated likelihood and prior assumptions, and $\boldsymbol{G}_{:,k}$
is the $k$th column of $\boldsymbol{G}$. Usually, the minimization
problem (6) is solved under the additional orthonormality constraint
\begin{equation}
\boldsymbol{G}\boldsymbol{G}^{T}=\boldsymbol{I}
\end{equation}
This constraint is imposed so as to ensure non-degeneracy, i.e., to
prevent the bases in the factor loadings matrix $\boldsymbol{G}$
from becoming \emph{degenerate}. However, it is effective only in
cases of \emph{undercomplete} or \emph{complete} representations,
i.e., the number of latent features does \emph{not} exceed the number
of observed features ($K\leq D$) \cite{rica}. 

As we discussed in Section 1, the capacity of extracting overcomplete
latent feature representations is a significant merit for unsupervised
feature learning algorithms. As such, it is important that ICA can
be effectively employed when postulating $K>D$. A computationally
efficient method that resolves this issue was proposed in \cite{rica};
it consists in replacing the orthonormality constraint (7) with a
soft reconstruction cost which measures the difference between the
original observations $\{\boldsymbol{x}_{n}\}_{n=1}^{N}$ and the
reconstructions obtained by a linear autoencoder, where the encoding
and decoding weights are tied to the feature matrix $\boldsymbol{G}$
learned by the model. The resulting method, dubbed reconstruction
ICA (RICA), yields the minimization problem
\[
\underset{\boldsymbol{G}}{\mathrm{min}}\frac{\xi}{N}\sum_{n=1}^{N}||\boldsymbol{G}\boldsymbol{G}^{T}\boldsymbol{x}_{n}-\boldsymbol{x}_{n}||_{2}^{2}+\sum_{k=1}^{K}\sum_{n=1}^{N}h(\boldsymbol{G}_{:,k}^{T}\boldsymbol{x}_{n})
\]
where $\xi$ is a regularization parameter, and $\boldsymbol{G}_{:,k}$
is the $k$th column of $\boldsymbol{G}$.

As discussed in \cite{rica}, under this scheme some bases of $\boldsymbol{G}$
may still degenerate and become zero, because the reconstruction constraint
can be satisfied with only a complete subset of features\footnote{The authors of \cite{rica} resorted to introducing an additional
norm ball constraint to resolve this issue.}. In addition, we note that the imposed reconstruction error constraints
suffer from a weak point that has been extensively studied in the
autoencoder (AE) literature: specifically, the optimal reconstruction
criterion of AEs may merely lead to the trivial solution of just copying
the input to the output, that yields very low reconstruction error
in a given training set combined with extremely poor modeling and
generalization performance \cite{ae-issues} (performing training
using a \emph{noise-corrupted} version of the original observations
is a solution commonly used in AE literature to prevent this from
happening \cite{ae-issues}).

\section{Proposed Approach}

\subsection{IBP-ICA}

\subsubsection{Model Formulation}

Let us consider a set of $D$-dimensional observations $\{\boldsymbol{x}_{n}\}_{n=1}^{N}$.
We model this dataset using ICA, adopting the conventional assumptions
(1) - (5). However, in contrast to the conventional model formulation,
we specifically want to examine the case where the dimensionality
$K$ of the latent feature vectors $\boldsymbol{y}_{n}$ tends to
infinity, $K\rightarrow\infty$. In other words, we seek to obtain
a nonparametric formulation for our model. 

Under such an assumption, and imposing an appropriate prior distribution
over the latent feature matrix $\boldsymbol{G}$, we can obtain an
inference algorithm that allows for automatic determination of the
most appropriate number of latent features to model our data, and
performs inference over only this finite set \cite{ibp2}. For this
purpose, we impose a spike-and-slab prior over the components of the
\emph{latent feature matrix }$\boldsymbol{G}$:
\begin{equation}
p(g_{dk}|z_{dk};\lambda_{k})=z_{dk}\mathcal{N}(g_{dk}|0,\lambda_{k}^{-1})+(1-z_{dk})\delta_{0}(g_{dk})
\end{equation}
where $\lambda_{k}$ is the precision parameter of the (Gaussian)
prior distribution of the $k$th base in $\boldsymbol{G}$, $\delta_{0}(\cdot)$
is a \emph{spike} distribution with all its mass concentrated at zero
(delta function), and the discrete latent variables $z_{dk}$ indicate\emph{
latent feature activity, }being equal to one if the $k$th latent
feature contributes to generation of the $d$th observed dimension
(i.e., the latent feature is \emph{active}), zero otherwise. 
Note that a similar prior has been previously adopted in the related, 
factor analysis (FA)-based latent feature model of \cite{related}. 
The \emph{key} difference between FA- and ICA-based models 
is that in FA the prior over the latent feature vectors is a spherical 
Gaussian; in contrast, in ICA  we impose independent priors over 
each latent feature taking the form of a more complex distribution 
(a Gaussian mixture in our work, see Eqs. (2)-(4)).

Spike-and-slab priors \cite{ssl} are commonly used to introduce \emph{sparsity
}in the modeling procedure; combined with a nonparametric prior over
the matrix of discrete latent variables $\boldsymbol{Z}=[z_{dk}]_{d,k}$,
they also allow for defining a generative process for the number of
latent factors under a sparse modeling scheme. To this end, we utilize
the IBP prior \cite{ibp}; specifically, we adopt the \emph{stick-breaking}
construction of IBP \cite{ibp-sb-vb}. This is another \emph{key} difference
between our model formulation and the method of \cite{related}; the
major advantage of using the \emph{stick-breaking} construction consists
in allowing for obtaining a variational inference algorithm, which
is much more scalable to massive data compared to MCMC \cite{ibp-sb-vb}
(used in \cite{related}). We have: 
\begin{equation}
z_{dk}\sim\mathrm{Bernoulli}(\pi_{k})
\end{equation}
 where
\begin{equation}
\pi_{k}=\prod_{i=1}^{k}v_{i}
\end{equation}
 and the prior over the stick-variables $v_{i}$ is defined as
\begin{equation}
v_{k}\sim\mathrm{Beta}(\alpha,1)
\end{equation}
 In (11), $\alpha$ is called the innovation hyperparameter, and controls
the tendency of the process to discover new latent features. We impose
a Gamma hyperprior over it, yielding
\begin{equation}
p(\alpha)=\mathcal{G}(\alpha|\gamma_{1},\gamma_{2})
\end{equation}
 Finally, we impose a Gamma prior over the precision parameters $\lambda_{k}$,
which reads
\begin{equation}
p(\lambda_{k})=\mathcal{G}(\lambda_{k}|c,f)
\end{equation}
 as well as a Gamma prior over the noise precision parameter $\phi$
\begin{equation}
p(\phi)=\mathcal{G}(\phi|a,b)
\end{equation}
 This concludes the definition of our IBP-ICA model.

\subsubsection{Inference Algorithm}

The formulation of our model using the stick-breaking construction
of IBP allows for performing inference by means of an efficient hybrid
variational algorithm, inspired from \cite{hvb}. Our approach combines:
(i) \emph{mean-field} variational inference \cite{vbg} for the model
parameters and latent variables, similar to existing models utilizing
the stick-breaking construction of the IBP, e.g. \cite{ibp-sb-vb};
and (ii) a \emph{local} Metropolis-Hastings (MH) step to sample from
the distribution over the number of (active) latent features pertaining
to each dimension of the observed data, inspired from \cite{ibp2}.

Let us denote as $q(\cdot)$ the obtained variational posteriors.
Following \cite{ibp2}, the proposed number of new features $K_{d}^{*}$
to be added to the number of active features $K_{d}$ pertaining to
the $d$th input dimension is sampled from the Poisson proposal distribution:
\begin{equation}
p(K_{d}^{*})=\mathrm{Poisson}\left(K_{d}^{*}\big|\frac{\mathbb{E}_{q(\alpha)}[\alpha]}{D-1}\right)
\end{equation}
 This proposal is accepted with probability
\begin{equation}
p_{d}^{*}=\mathrm{min}\left\{ 1,\frac{\theta_{d}^{*}}{\theta_{d}}\right\} 
\end{equation}
 where \cite{ibp2}: 
\begin{equation}
\theta_{d}^{*}=|\boldsymbol{M}_{d}^{*}|^{-N/2}\mathrm{exp}\left(\frac{1}{2}\sum_{n=1}^{N}\boldsymbol{m}_{nd}^{*T}\boldsymbol{M}_{d}^{*}\boldsymbol{m}_{nd}^{*}\right)
\end{equation}
\begin{equation}
\begin{aligned}\boldsymbol{m}_{nd}^{*}= & \mathbb{E}_{q(\phi)}[\phi]\boldsymbol{M}_{d}^{*-1}\boldsymbol{G}_{d,:}^{*T}\\
 & \times\left(x_{nd}-\mathbb{E}_{q(\boldsymbol{y}_{n}).q(\boldsymbol{G})}[\boldsymbol{G}_{d,:}\boldsymbol{y}_{n}]\right)
\end{aligned}
\end{equation}
\begin{equation}
\boldsymbol{M}_{d}^{*}=\mathbb{E}_{q(\phi)}[\phi]\boldsymbol{G}_{d,:}^{*T}\boldsymbol{G}_{d,:}^{*}+\boldsymbol{I}
\end{equation}
 the $1\times K_{d}^{*}$ variables $\boldsymbol{G}_{d,:}^{*}$ in
(18) and (19) are sampled from their prior (8), and
\begin{equation}
\theta_{d}=|\boldsymbol{M}_{d}|^{-N/2}\mathrm{exp}\left(\frac{1}{2}\sum_{n=1}^{N}\boldsymbol{m}_{nd}^{T}\boldsymbol{M}_{d}\boldsymbol{m}_{nd}\right)
\end{equation}
\begin{equation}
\begin{aligned}\boldsymbol{m}_{nd}= & \mathbb{E}_{q(\phi)}[\phi]\boldsymbol{M}_{d}^{-1}\mathbb{E}_{q(\boldsymbol{G})}[\boldsymbol{G}_{d,:}^{T}]\\
 & \times\left(x_{nd}-\mathbb{E}_{q(\boldsymbol{y}_{n}).q(\boldsymbol{G})}[\boldsymbol{G}_{d,:}\boldsymbol{y}_{n}]\right)
\end{aligned}
\end{equation}
\begin{equation}
\boldsymbol{M}_{d}=\mathbb{E}_{q(\phi)}[\phi]\mathbb{E}_{q(\boldsymbol{G})}[\boldsymbol{G}_{d,:}^{T}\boldsymbol{G}_{d,:}]+\boldsymbol{I}
\end{equation}
(the expressions of the expectations $\mathbb{E}_{q}[\cdot]$ 
can be straightforwardly derived by following the identities
pertaining to Gaussians, Gamma, and Beta distributions in 
\cite[Appendix B]{BishopBook}). Note that this MH step is 
of a \emph{local} (input dimension-wise) nature, and
thus is very fast. Further, the introduction of this local sampling
step is the reason why our algorithm does not require provision of
heuristic truncation thresholds, because MCMC samplers for nonparametric
Bayesian models can operate in an unbounded feature space \cite{mc2}. 

Having updated the number of latent features, our inference algorithm
proceeds to obtain the variational posteriors over the latent feature
activity indicator variables $z_{dk}$. This is performed by maximization
of the variational free energy of the model over $q(z_{dk}=1)$, yielding
\begin{equation}
q(z_{dk}=1)=\frac{1}{1+\mathrm{exp}(-\omega_{dk})}
\end{equation}
 where
\begin{equation}
\begin{aligned}\omega_{dk}= & \sum_{i=1}^{k}\mathbb{E}_{q(v_{i})}[\mathrm{log}v_{i}]+\mathbb{E}_{q(\boldsymbol{v})}\left[\mathrm{log}\left(1-\prod_{i=1}^{k}v_{i}\right)\right]\\
 & +\mathbb{E}_{q(\lambda_{k}),q(g_{dk})}[\mathrm{log}\mathcal{N}(g_{dk}|0,\lambda_{k}^{-1})]
\end{aligned}
\end{equation}

Further, the posterior over the latent feature matrices is obtained
by maximization of the variational free energy over $q(g_{dk})$,
yielding
\begin{equation}
\begin{aligned}q(g_{dk})= & q(z_{dk}=1)\mathcal{N}(g_{dk}|\tilde{\mu}_{dk},\tilde{\lambda}_{dk}^{-1})\\
 & +\left(1-q(z_{dk}=1)\right)\delta_{0}(g_{dk})
\end{aligned}
\end{equation}
where
\begin{equation}
\tilde{\lambda}_{dk}=\mathbb{E}_{q(\phi)}[\phi]\sum_{n=1}^{N}\mathbb{E}_{q(y_{nk})}[y_{nk}^{2}]+\mathbb{E}_{q(\lambda_{k})}[\lambda_{k}]\;\forall d
\end{equation}
\begin{equation}
\begin{aligned}\tilde{\mu}_{dk}=\tilde{\lambda}_{dk}^{-1}\mathbb{E}_{q(\phi)}[\phi]\sum_{n=1}^{N} & \mathbb{E}_{q(y_{nk})}[y_{nk}]\\
 & \times\left(x_{nd}-\mathbb{E}_{q(\boldsymbol{y}_{n}),q(\boldsymbol{G})}[\boldsymbol{G}_{d,:}\boldsymbol{y}_{n}]\right)
\end{aligned}
\end{equation}

Similarly, the posterior over the learned latent features yields
\begin{equation}
q(y_{nk})=\mathcal{N}(y_{nk}|\tilde{m}_{nk},\tilde{s}_{nk})
\end{equation}
 where
\begin{equation}
\tilde{s}_{nk}^{-1}=\mathbb{E}_{q(\phi)}[\phi]\mathbb{E}_{q(\boldsymbol{G})}[\boldsymbol{G}_{:,k}^{T}\boldsymbol{G}_{:,k}]+\sum_{j=1}^{J}\zeta_{nkj}\mathbb{E}_{q(s_{kj})}[s_{kj}^{-1}]
\end{equation}
\begin{equation}
\tilde{m}_{nk}=\tilde{s}_{nk}\mathbb{E}_{q(\phi)}[\phi]\mathbb{E}_{q(\boldsymbol{G})}[\boldsymbol{G}_{:,k}^{T}]\boldsymbol{x}_{n}
\end{equation}
In the above expressions, $\zeta_{nkj}$ are the source model component
posteriors, which read
\begin{equation}
\zeta_{nkj}\propto\mathrm{exp}\big(\mathbb{E}_{q(\boldsymbol{\varpi}_{k})}[\mathrm{log}\varpi_{kj}]+\mathbb{E}_{q(s_{kj})}[\mathrm{log}\mathcal{N}(y_{nk}|0,s_{kj})]\big)
\end{equation}
 where the posterior over the mixture component weights yields

\begin{equation}
q(\boldsymbol{\varpi}_{k}|\tilde{\boldsymbol{\xi}}_{k})=\mathrm{Dir}(\boldsymbol{\varpi}_{k}|\tilde{\boldsymbol{\xi}}_{k})
\end{equation}
 with $\tilde{\boldsymbol{\xi}}_{k}=[\tilde{\xi}_{kj}]_{j=1}^{J}$,
and 
\begin{equation}
\tilde{\xi}_{kj}=\xi_{kj}+\sum_{n=1}^{N}\zeta_{nkj}
\end{equation}
 while the posterior over the $s_{kj}$ obtains
\begin{equation}
q(s_{kj}^{-1})=\mathcal{G}(s_{kj}^{-1}|\tilde{\eta}_{kj1},\tilde{\eta}_{kj2})
\end{equation}
 where
\begin{equation}
\tilde{\eta}_{kj1}=\eta_{1}+\frac{1}{2}\sum_{n=1}^{N}\zeta_{nkj}
\end{equation}
 and
\begin{equation}
\tilde{\eta}_{kj2}=\eta_{2}+\frac{1}{2}\sum_{n=1}^{N}\zeta_{nkj}\mathbb{E}_{q(y_{nk})}[y_{nk}^{2}]
\end{equation}

The rest of the model posteriors take expressions identical to existing
variational inference algorithms for nonparametric Bayesian formulations
of ICA, e.g., \cite{ibp-sb-vb}. For completeness sake, we provide
these expressions in the Appendix. An outline of the inference algorithm
of IBP-ICA is provided in Alg. 1.

\begin{algorithm}
\protect\caption{IBP-ICA inference algorithm. }

\begin{enumerate}
\item Select the number of source model components $J$, as well as the
hyperparameters of the imposed model priors: $a,b,\gamma_{1},\gamma_{2},c,f$,
$\eta_{1}$, $\eta_{2}$, and $\xi_{kj},\;\forall k,j$.
\item For $MAXITER$ iterations or until convergence of the variational
free energy of the model, \textbf{do}:\end{enumerate}
\begin{itemize}
\item Perform the MH step (15)-(16) to update the number of generated latent
features.
\item Update the variational posterior over the latent feature activity
variables, $q(\boldsymbol{Z})$, using (23).
\item Update the posteriors $q(y_{nk})$, $q(\boldsymbol{\varpi}_{k})$,
$q(s_{kj}^{-1})$, $\forall n,k,j$, using (28)-(36).
\item Update the posteriors $q(\lambda_{k})$, $q(v_{k})$, $q(\phi)$,
$q(\alpha)$, $\forall k$, using the expressions given in the Appendix.
\item Update the posteriors $q(\boldsymbol{G})=\{q(g_{dk})\}_{d,k}$ using
(25)-(27).\end{itemize}
\end{algorithm}

\begin{figure}

\centering{}%
\includegraphics[scale=0.34]{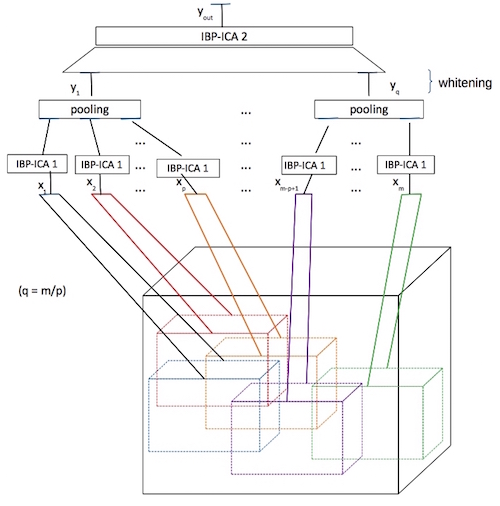}\protect\caption{The proposed SC-IBP-ICA architecture. It is trained by using some
randomly selected video patches, from each of which we extract an
input vector $\boldsymbol{x}$. Pooling is performed between the first
and the second layer over small neighborhoods comprising $p$ adjacent
first layer units. The outputs of the first layer are presented as
inputs to the second layer (after whitening).}

\end{figure}

\subsection{Stacked Convolutional IBP-ICA Networks}

Let us now turn to the problem of action recognition in video sequences.
Since videos are sequences of images (frames), to address this task
using our model we simply extract the component frames of the video
sequences at hand and flatten them into a vector, which is eventually
presented to our model as the observed input. However, a problem with
this model setup is the resulting extremely high dimensionality of
the input observations, which induces an excessive increase to the
number of inferred model parameters, and, hence, the computational
complexity of the inference algorithm of our model.

To alleviate this computational burden, we opt for resorting to a
\emph{stacked convolutional IBP-ICA network}
architecture, inspired from local receptive field networks 
(e.g., \cite{isa,cdbn}). Specifically, instead of processing the 
whole video frames using a single IBP-ICA model,
we use multiple convolved copies of an IBP-ICA model to process smaller
(overlapping) patches of the video frame sequences. This way, we manage
to dramatically reduce the number of inferred model parameters, and,
hence, the imposed computational complexity. The output feature vectors
of the postulated component IBP-ICA models are further processed by
a pooling sublayer, similar to conventional
convolutional neural network architectures. The introduction of pooling
into our convolutional IBP-ICA network endows it with the merit of
translational invariance, which is a desideratum in the context of
action recognition applications in video sequences. Finally, we stack
multiple layers of our convolutional IBP-ICA network to obtain a deep
learning architecture. Our deep architecture allows for capturing
features that correspond to both lower and higher level analysis of
the spatiotemporal dynamics in the observed data; this way, it extracts
much richer information to train a classifier with, compared to shallow
models \cite{sae} .

Training of our proposed SC-IBP-ICA network proceeds as follows: Initially,
we train an IBP-ICA model on \emph{small video patches}. Subsequently,
we build a \emph{convolutional network} of IBP-ICA models, by \emph{replicating}
and applying copies of the learned IBP-CA model to\emph{ different
overlapping patches }of the input video frame sequences (with an additional
\emph{pooling} sublayer on top). Finally, we feed the output of our
convolutional IBP-ICA network to a similarly trained \emph{subsequent}
convolutional IBP-ICA network, thus creating a deep learning architecture
through \emph{stacking}\footnote{In our experiments, input data are additionally pre-processed and
whitened using PCA, exactly as described in \cite{isa,rica}.}. As a final note, we underline that training of a whole SC-IBP-ICA
network is performed in a greedy layerwise manner, similar to many
existing convolutional deep learning architectures (e.g., \cite{cdbn}).

A graphical illustration of the proposed SC-IBP-ICA network and its
training procedures is provided in Fig. 1.

\subsection{Feature Generation Using SC-IBP-ICA}

Given a learned SC-IBP-ICA network, feature generation from observed
video frame sequences is performed similar to conventional deep convolutional
networks, by means of feedforward computation. Specifically, based
on the expression of the variational posterior $q(y_{nk})$ derived
in Eq. (28), to perform \emph{feature generation} we feedforward the
input vectors $\boldsymbol{x}_{n}$ by application of Eq. (30), using
the \emph{already} \emph{obtained }values of $\mathbb{E}_{q(\phi)}[\phi]$,
$\mathbb{E}_{q(\boldsymbol{G})}[\boldsymbol{G}_{:,k}^{T}\boldsymbol{G}_{:,k}]$,
and $\mathbb{E}_{q(\boldsymbol{G})}[\boldsymbol{G}_{:,k}^{T}]$. Note
that, from the simple feedforward computation form of Eq. (30), it
directly follows that using our method to generate features is extremely
fast, with costs identical to existing state-of-the-art approaches,
e.g. \cite{isa}.

\begin{table}
\protect\caption{Mean average precision on the Hollywood2 dataset.}

\centering{}%
\begin{tabular}{|c|c|}
\hline 
Algorithm & Average Precision\tabularnewline
\hline 
\hline 
Harris3D + HOG/HOF \cite{pipeline} & 45.2\%\tabularnewline
\hline 
Hessian + ESURF \cite{pipeline} & 38.2\%\tabularnewline
\hline 
Cuboids + HOG/HOF \cite{pipeline} & 46.2\%\tabularnewline
\hline 
convGRBM \cite{grbm} & 46.6\%\tabularnewline
\hline 
Dense + HOG3D \cite{pipeline} & 45.3\%\tabularnewline
\hline 
ISA (with reconstruction & \multirow{2}{*}{54.6\%}\tabularnewline
penalty) \cite{rica} & \tabularnewline
\hline 
SC-IBP-ICA (1 layer) & 47.8\% (286)\tabularnewline
\hline 
SC-IBP-ICA (2 layers) & 53.5\% (286 - 198)\tabularnewline
\hline 
\end{tabular}
\end{table}

\begin{table}
\protect\caption{Average accuracy on the YouTube dataset.}

\centering{}%
\begin{tabular}{|c|c|}
\hline 
Algorithm & Accuracy\tabularnewline
\hline 
\hline 
HAR + HES  & \multirow{2}{*}{71.2\%}\tabularnewline
+ MSER + SIFT \cite{pipeline2} & \tabularnewline
\hline 
Harris3D + Grads.  & \multirow{2}{*}{71.2\%}\tabularnewline
+ PCA + Heuristics \cite{pipeline2} & \tabularnewline
\hline 
ISA \cite{isa} & 75.8\%\tabularnewline
\hline 
SC-IBP-ICA (1 layer) & 73.6\% (291.7)\tabularnewline
\hline 
SC-IBP-ICA (2 layers) & 75.4\% (291.7 - 197.9)\tabularnewline
\hline 
\end{tabular}
\end{table}

\begin{table}
\protect\caption{Average accuracy on the KTH dataset.}

\centering{}%
\begin{tabular}{|c|c|}
\hline 
Algorithm & Accuracy\tabularnewline
\hline 
\hline 
Harris3D + HOG/HOF \cite{pipeline} & 91.8\%\tabularnewline
\hline 
Hessian + ESURF \cite{pipeline} & 81.4\%\tabularnewline
\hline 
Cuboids + HOG3D \cite{pipeline} & 90.0\%\tabularnewline
\hline 
HMAX \cite{jhuang} & 91.7\%\tabularnewline
\hline 
3D-CNN \cite{3dcnn} & 90.2\%\tabularnewline
\hline 
convGRBM \cite{grbm} & 90.0\%\tabularnewline
\hline 
ISA \cite{isa} & 93.9\%\tabularnewline
\hline 
SC-IBP-ICA (1 layer) & 92.3\% (293)\tabularnewline
\hline 
SC-IBP-ICA (2 layers) & 93.4\% (293 - 195)\tabularnewline
\hline 
\end{tabular}
\end{table}

\begin{table}
\protect\caption{Average feature extraction time in our experiments (Hollywood2 dataset).}

\centering{}%
\begin{tabular}{|c|c|}
\hline 
Algorithm & Seconds/Frame\tabularnewline
\hline 
\hline 
HOG3D & 0.20\tabularnewline
\hline 
ISA \cite{isa} (1 layer)  & 0.13\tabularnewline
\hline 
ISA \cite{isa} (2 layers)  & 0.40\tabularnewline
\hline 
SC-IBP-ICA (1 layer) & 0.12\tabularnewline
\hline 
SC-IBP-ICA (2 layers) & 0.38\tabularnewline
\hline 
\end{tabular}
\end{table}

\section{Experiments}

In this section, we experimentally investigate how SC-IBP-ICA compares
to the current state-of-the-art in action recognition. To perform
our experimental investigations, we use three publicly available action
recognition benchmarks, namely Hollywood2 \cite{hollywood}, KTH actions,
and YouTube actions \cite{pipeline2}. Our experimental setup 
is the same as in \cite{isa,rica}, adopting exacty the same data 
preprocessing/postprocessing steps. After extracting local features 
by means of SC-IBP-ICA, we subsequently perform \emph{vector quantization} of
the obtained feature vectors using K-means. Finally, we use these
discretized feature vectors to train an SVM classifier \cite{svm-book} 
employing a $\chi^{2}$ kernel. 

We adopt the same dataset splits and evaluation metrics as in \cite{pipeline,pipeline2}.
Specifically, Hollywood2 human actions dataset contains 823 train
and 872 test video clips organized into 12 action classes; each video
clip may have more than one action label. We utilize the produced
feature vectors to train 12 binary SVM classifiers, one for each action.
We use the final average precision (AP) metric for our evaluations,
computed as the average of AP for each classifier run on the test
set. Youtube actions dataset contains 1600 video clips organized into
11 action classes. These video clips have been split into 25 folds
which we use to perform 25-fold cross-validation. Note that, from
each split, we use only videos indexed 01 to 04, except for the biking
and walking classes, where we use the whole datasets. Finally, KTH
actions dataset contains 2391 video samples organized into 6 action
classes. We split these samples into a test set containing subjects
2, 3, 5, 6, 7, 8, 9, 10, 12, and a training set containing the rest
of the subjects. We use the produced feature vectors to train a multi-class
SVM.

We evaluate SC-IBP-ICA architectures comprising one and two layers,
to examine how extra layer addition affects model performance. Regarding
selection of the size of receptive fields of our model, \emph{the
first layer is of size 16$\times$16 (spatial) and 10 (temporal),
while the second one is of size 20$\times$20 (spatial) and 14 (temporal)},
similar to \cite{isa,pipeline}. The size of the output of the pooling
layers is identical to the size of the output of the IBP-ICA models
that feed it, i.e. the number of latent features our method discovers.
Training is performed on 200,000 video blocks, randomly sampled from the training
set of each dataset. We perform dense sampling with \emph{50\% overlap
in all dimensions}. In cases of 2-layer architectures, we train the
used SVM classifiers by combining the features generated from \emph{both}
layers. This setup retains more representative features compared to
using only the features from the \emph{top} layer, corresponding to
a coarse-to-fine analysis of the observations \cite{cdbn}. 

Our obtained results are provided in Tables 1-3. Note that the performances
of the competing methods reported therein have been cited from \cite{isa} 
and \cite{rica}. The reported results of ISA were obtained with 300 latent features on the first
layer, and 200 latent features on the second layer (these have been
heuristically found to yield the best performance
among a large set of evaluated alternatives).

In Tables 1-3, we also provide the number of latent features \emph{automatically}
discovered by our method (in parentheses, beside the accuracy figures).
Note that the reported performance results and the corresponding numbers
of discovered latent features pertaining to the YouTube dataset, illustrated
in Table 2, are means over the 25 splits of the dataset into training
and test sets (folds) provided by its creators. We observe that our
method obtains performance similar to the state-of-the-art, while
also allowing for automatic inference of the appropriate number of
generated features. We also observe that addition of a second layer
is auspicious in all cases, corroborating similar findings in the
literature.

In Table 4, we depict the computational costs of our approach regarding
feature extraction from the Hollywood2 (test) dataset (run as a single
thread)\footnote{We run these experiments on an Intel Xeon 2.5GHz Quad-Core CPU with
64GB RAM. Our source codes were written in MATLAB R2014a.}. It is clear that feature generation using our method takes time
similar to existing ICA-based approaches, namely the ISA method presented
in \cite{isa}, as theoretically expected (the small computational
advantage of our method is presumably due to the lower number of latent
features compared to \cite{isa}).

Further, we examine model generalization performance under a \emph{transfer
learning} setting: In real-world settings, a feature extraction system
pre-trained on samples from a set of video sequences will be expected
to perform well on any previously unseen input video sequence, with
\emph{no} samples of it included in its training set. To perform this
kind of evaluations, we train our SC-IBP-ICA network on video blocks
randomly sampled from the KTH dataset, and evaluate its performance
on the Hollywood2 dataset. Under this setup, our method yields a mean
average precision equal to 51.9\%. Compare this result to the performance
obtained by ISA \cite{isa} under the same experimental setup, which
yields a mean average precision equal to 50.8\%.

Finally, an interesting question concerns how IBP-ICA model performance
changes in case we use cross-validation to perform model selection
instead of sampling from the related posteriors over the number of
features {[}Eqs. (15)-(22){]}. To investigate this, we repeat our
experiments using the Hollywood2 dataset in the following way: We
perform model training without sampling the number of features, which
is considered a given constant. We repeat this experiment multiple
times, with different numbers of features each time; we try configurations
comprising 250-350 features on the first layer, and 150-250 features
on the second layer, with a step of 5 features between consecutive
evaluated models. Model selection is performed on the grounds of the
accuracy obtained in the available test set.

Our findings are illustrated in Table 5; as we observe, cross-validation
yields a slightly better model performance than our fully-fledged
nonparametric Bayesian approach. However, these mediocre gains come
at the price of significant computational costs: Specifically, the
computational gain from skipping the updates of the posterior over
the number of latent features constitutes only a meager 16.2\% of
the total training time. On the other hand, the aforementioned cross-validation
procedure required evaluating 40 different model configurations, i.e.
repeating model training 40 times. Therefore, model selection by means
of our nonparametric Bayesian approach offers an overwhelmingly favorable
complexity/accuracy trade-off compared to an exhaustive cross-validation
technique.

In the same vein, another interesting question concerns comparison
of the proposed hybrid variational inference algorithm of our model
with the straightforward alternative of Markov chain Monte-Carlo (MCMC)
inference. To examine this aspect, we rerun our experiments by properly
adapting the MCMC algorithm outlined in \cite{related} in the context
of our model. As we observed, our proposed algorithm requires one
order of magnitude less time to converge, for a negligible
performance deterioration compared to MCMC.

\begin{table}
\protect\caption{Mean average precision on the Hollywood2 dataset by application of
cross-validation.}

\centering{}%
\begin{tabular}{|c|c|c|}
\hline 
Algorithm & Average Precision & Model Size\tabularnewline
\hline 
\hline 
SC-IBP-ICA (1 layer) & 48.1\% & 295\tabularnewline
\hline 
SC-IBP-ICA (2 layers) & 53.9\% & 295-205\tabularnewline
\hline 
\end{tabular}
\end{table}

\section{Conclusions}

In this paper, we introduced a deep convolutional nonparametric Bayesian
approach for unsupervised feature extraction. The main building block
of our approach is a nonparametric Bayesian formulation of ICA, dubbed
IBP-ICA. Our method imposes a spike-and-slab prior over the factor
loadings matrices, driven by an IBP prior over the latent feature
activity indicators. This way, it allows for automatic \emph{data-driven}
inference of the most appropriate number of latent features. This
is in \emph{stark} contrast with all existing methods, such as DBNs
\cite{dbn}, ICA variants \cite{isa,rica}, and SAEs \cite{sae},
where hand-tuning the number of extracted latent features is an \emph{essential}
part of the application of these methods to real-life tasks. It is also 
substantially different from the related approach of \cite{betaprocess}, where, instead of the
spike-and-slab prior used in this work, a simpler Beta-Bernoulli process 
prior is employed; this formulation of \cite{betaprocess} 
does \emph{not} allow for performing feature generation via 
\emph{simple} feedforward computation [in the sense of Eq. (30)].
Hence, feature generation in \cite{betaprocess} requires much higher 
computational costs compared to state-of-the-art deep learning
approaches and our method. In addition, our approach can model latent feature 
distributions of arbitrary complexity (approximated via
mixtures of Gaussians, Eq. (2)), as opposed to \cite{betaprocess} 
which postulates a simplistic spherical Gaussian prior.

We devised an efficient variational inference algorithm for our model.
Our method is very easy to train because
(batch) variational inference does not need any tweaking with heuristics
such learning rates and convergence criteria. In this regard, our
method lies on exactly the opposite side of the spectrum compared
to conventional approaches based on neural networks: Our method entails
\emph{no }need of selection of training algorithm heuristics, whatsoever,
while training neural networks is a tedious procedure requiring a
great deal of hand-tuning of several heuristics (e.g., learning
rate, weight decay, convergence parameters, inertia).

We evaluated our approach using three well-known action recognition
benchmarks, adopting a standard video processing pipeline (e.g., \cite{neurosc2}).
As we showed, our method yields results similar to the state-of-the-art
for these benchmarks, while imposing competitive computational costs
for \emph{feature generation}. These results corroborate that 
nonparametric Bayesian models can offer a viable alternative
to existing deep feature extractors, and at the same time mitigate
some of the major hurdles deep nets are confronted with, regarding
data-driven selection of model size during inference, and learning
algorithm parameters fine-tuning.

\section*{Appendix}

We have \begin{equation}
q(\lambda_{k})=\mathcal{G}(\lambda_{k}|\tilde{c}_{k},\tilde{f}_{k})
\end{equation}
 where
\begin{equation}
\tilde{c}_{k}=c+\frac{1}{2}\sum_{d=1}^{D}q(z_{dk}=1),\;\;\tilde{f}_{k}=f+\sum_{d=1}^{D}\mathbb{E}_{q(g_{dk})}[g_{dk}^{2}]
\end{equation}
and \begin{equation}
\begin{aligned}q(\phi)=\mathcal{G}\big(\phi\big|a+\frac{ND}{2},b+ & \sum_{n=1}^{N}(\boldsymbol{x}_{n}-\mathbb{E}_{q(\boldsymbol{y}_{n}),q(\boldsymbol{G})}[\boldsymbol{G}\boldsymbol{y}_{n}])^{T}\\
 & \times(\boldsymbol{x}_{n}-\mathbb{E}_{q(\boldsymbol{y}_{n}),q(\boldsymbol{G})}[\boldsymbol{G}\boldsymbol{y}_{n}])\big)
\end{aligned}
\end{equation}
For the stick-variables $v_{k}$, we adopt the approximations \cite{ibp-sb-vb}: 
\begin{equation}
q(v_{k})=\mathrm{Beta}(v_{k}|\tilde{\tau}_{k},\hat{\tau}_{k})
\end{equation}
\begin{equation}
\begin{aligned}\tilde{\tau}_{k}= & \sum_{m=k+1}^{K}\left(D-\sum_{d=1}^{D}q(z_{dm}=1)\right)\sum_{i=k+1}^{m}q_{i}\\
 & +\sum_{d=1}^{D}\sum_{m=k}^{K}q(z_{dm}=1)+\mathbb{E}_{q(\alpha)}[\alpha]
\end{aligned}
\end{equation}
\begin{equation}
\hat{\tau}_{k}=1+\sum_{m=k}^{K}\left(D-\sum_{d=1}^{D}q(z_{dm}=1)\right)q_{k}
\end{equation}
 where we denote $K\triangleq\underset{d}{\mathrm{max}}\;K_{d}$,
\begin{equation}
\begin{aligned}q_{k}\propto\mathrm{exp} & \big(\psi(\hat{\tau}_{k})+\sum_{i=1}^{k-1}\psi(\tilde{\tau}_{i})-\sum_{i=1}^{k}\psi(\tilde{\tau}_{i}+\hat{\tau}_{i})\big)\end{aligned}
\end{equation}
 and $\psi(\cdot)$ is the Digamma function. Finally, the innovation
hyperparameter yields: $q(\alpha)=\mathcal{G}(\alpha|\tilde{\gamma}_{1},\tilde{\gamma}_{2})$,
where $\tilde{\gamma}_{1}=\gamma_{1}+K-1$ and $\tilde{\gamma}_{2}=\gamma_{2}-\sum_{k=1}^{K-1}[\psi(\tilde{\tau}_{k})-\psi(\tilde{\tau}_{k}+\hat{\tau}_{k})]$.

\small

\bibliographystyle{ieee}
\bibliography{TMFA}

\end{document}